\documentclass{article}

\usepackage{arxiv}

\usepackage[utf8]{inputenc} 
\usepackage[T1]{fontenc}    
\usepackage{hyperref}       
\usepackage{url}            
\usepackage{booktabs}       
\usepackage{amsfonts}       
\usepackage{nicefrac}       
\usepackage{microtype}      
\usepackage{lipsum}
\usepackage{subcaption}
\usepackage{graphicx}
\usepackage{mathptmx} 
\usepackage{times} 
\usepackage{amsmath} 
\usepackage{amssymb}  
\graphicspath{ {./images/} }

\title{Sustainable Transfer Learning for Adaptive Robot Skills}

\author{
 Khalil Abuibaid \\
  Chair of Machine Tools and Control System,\\
  RPTU University Kaiserslautern-Landau,\\
  67663 Kaiserslautern,  Germany \\
  \texttt{khalil.abuibaid@rptu.de} \\
   \And
 Vinit Hegiste \\
  Chair of Machine Tools and Control System,\\
  RPTU University Kaiserslautern-Landau,\\
  67663 Kaiserslautern,  Germany \\
  \texttt{vinit.hegiste@rptu.de} \\
   \And
    Nigora Gafur \\
  Chair of Machine Tools and Control System,\\
  RPTU University Kaiserslautern-Landau,\\
  67663 Kaiserslautern,  Germany \\
  \texttt{nigora.gafur@rptu.de} \\
   \And
 Achim Wagner \\
  Innovative Factory Systems,\\
  German Institute of Artificial Intelligence,\\
  67663 Kaiserslautern, Germany,\\
  \texttt{achim.wagner@dfki.de} \\
    \And
    Martin Ruskowski \\
Chair of Machine Tools and Control Systems, \\ 
RPTU University Kaiserslautern-Landau,  \\
67663 Kaiserslautern, Germany  \\
  and  \\
  Innovative Factory Systems,\\
  German Institute of Artificial Intelligence,\\
  67663 Kaiserslautern, Germany,\\
  \texttt{martin.ruskowski@dfki.de} \\
}

\begin{document}
\maketitle
\begin{abstract}
Learning robot skills from scratch is often time-consuming, while reusing data promotes sustainability, and improves sample efficiency. This study investigates policy transfer across different robotic platforms, focusing on peg-in-hole task using reinforcement learning (RL). Policy training is carried out on two different robots. Their policies are transferred and evaluated for zero-shot, fine-tuning, and training from scratch. Results indicate that zero-shot transfer leads to lower success rates and relatively longer task execution times while fine-tuning significantly improves performance with fewer training time-steps. These findings highlight that policy transfer with adaptation techniques improves sample efficiency and generalization, reducing the need for extensive retraining and supporting sustainable robotic learning. 
\end{abstract}

\section{Introduction}

The ability to transfer learned robot skills from one system to another is a crucial challenge in robotics. Reinforcement learning (RL) has revolutionized the field, significantly enhancing efficiency, reliability, and adaptability in robotic applications \cite{b1}. However, traditional RL approaches require extensive training time and large datasets for each robot and for each new task, making them computationally expensive and time-consuming for real-world applications. Addressing these challenges, the ability to reuse learned skills rather than retraining from scratch, has gained significant attention within the robotic and RL communities. Skill transfer is not only a critical step toward sustainable learning in robotics but also a key enabler of Industry 4.0 applications, where intelligent systems must seamlessly adapt to new environments.\\
In this paper, we present a use-case study comparing two learned policy models and investigating the possibility of skill transfer between robots performing the same task. Specifically, we focus on the peg-in-hole insertion task, where the learned policy must predict both the robot’s actions and the parameters of a hybrid force-motion controller. To achieve skill transfer, we explore policy adaptation through fine-tuning the policy neural network (NN) in a few-shot learning fashion. This approach offers a more sample-efficient alternative to training a robot from scratch, since RL methods often require millions of time-steps to achieve optimal policy. The remainder of this paper is organized as follows, Section \ref{sec:rel_Work} provides a brief overview of the methods used. Section \ref{sec:setup} describes the simulation setup, while Section \ref{sec:result} presents the results. Finally, we conclude the work with key findings and discuss how they set up for future work in Section \ref{sec:concolusion}.
\section{Related Work} \label{sec:rel_Work}
Skill transfer is crucial for ensuring adaptability across diverse robotic environments. The challenge of having variant domains leads to prolonged training times therefore it complicates the transfer learning process of a specific task. Recent advancements in learning from large datasets, as outlined in \cite{b2}, aggregate heterogeneous datasets collected from multiple robot platforms to improve policy generalization. This approach enables cross-embodiment skill transfer, allowing robots with different morphologies and control schemes to learn from shared experiences, thus reducing the need for extensive task-specific training. Another study \cite{b8} introduces a vision-based multi-robot dataset that improves policy generalization, enabling robots to adapt to new tasks with minimal additional training.\\
Early work on cross-embodiment transfer \cite{b5} explored adapting policies across different robotic simulations using a global policy model, incorporating variations in robot settings and kinematics. More recent studies \cite{b9} have integrated probabilistic meta-learning frameworks to adapt policies across different robot hardware, addressing variations in control schemes, camera perspectives, kinematics, and end-effector morphology. To tackle these challenges, \cite{b4} introduced a pipeline for training a single policy to operate across robots despite hardware differences, while other studies \cite{b7}, \cite{b11}, \cite{b3}, and \cite{b12} have explored generalist policy learning for multi-robot embodiments. Additionally, \cite{b13} proposes a shared modular policy to improve cross-platform generalization in robotic learning. \\
Transfer learning across robots has been extended through various approaches. Adaptive transfer methods encode state-action mappings across different domains to improve generalization \cite{b10}. Similarly, modular policy networks \cite{b6} enable multi-robot and multi-task skill transfer by decomposing policies into reusable components. Additionally, cycle generative networks have been utilized to learn the action space of another robot, facilitating skill adaptation across platforms \cite{b14}.
\section{Methodology} \label{sec:method}
\begin{figure}[hbt!]
    \centering
    \includegraphics[width=1.0\textwidth]{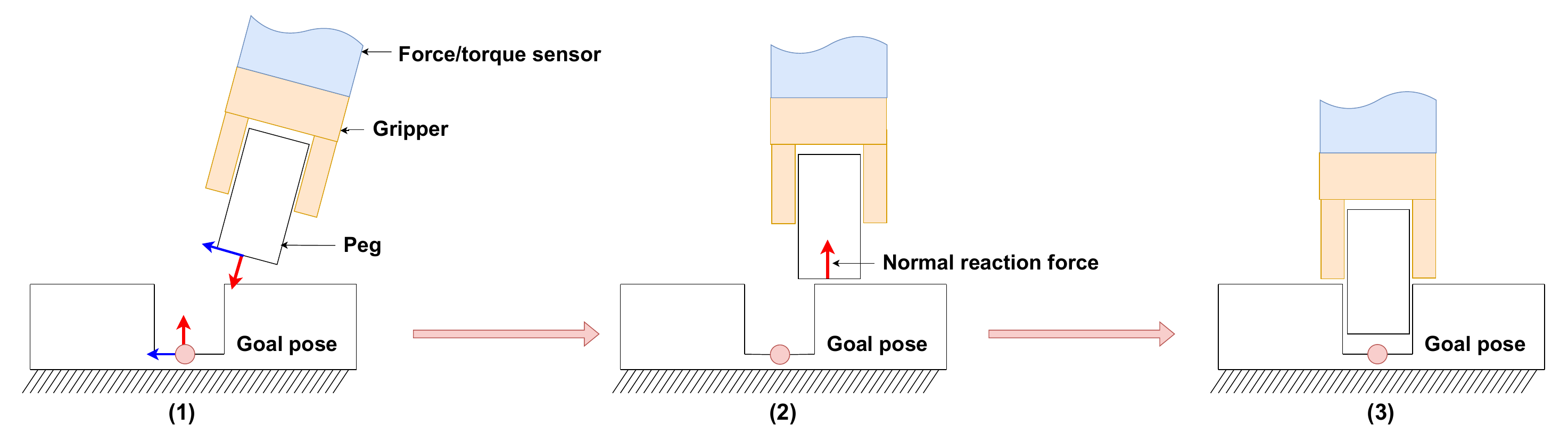}
    \caption{Ilustration of the peg in hole insertion task. }
    \label{fig:peginhole}
\end{figure}

This study presents a method for adaptive motion and force control to ensure precise performance in the task of peg in hole for position-controlled robots. Given the variability, RL helps explore and exploit the robot's dynamic behavior and define the optimal controller parameters accordingly. Therefore, the robots must determine the optimal path to find the hole as illustrated in Figure \ref{fig:peginhole}.
\subsection{Hybrid Motion and Force Control}
Hybrid motion-force control allows robots to simultaneously regulate motion and force, facilitating precise interaction tasks such as peg-in-hole insertion and assembly in unknown environments \cite{b15}. The controller is formulated mathematically as follows:
\begin{equation}
\begin{split}
    u_{hybrid}(t) = S(K_{p}^{x}x_{e} + K_{d}^{x}\dot{x}_{e})  + (I-S)(K_{p}^{f}F_{e}  +  K_{i}^{f
    } \int_{}^{} F_{e} dt)
    \end{split}
    \label{eq:hybrid}
\end{equation}
where $x_e$, $\dot{x}_e$ and $F_e$ are the position, velocity and force error. $S$ is a selection matrix that determines which degrees of freedom are controlled by motion, while the others are managed through force control. For example, $S$ can be determined as $  diag(\begin{bmatrix}1 & 1 & 0\end{bmatrix})$ where $ s_i \in [0,1]$,  indicating that motion control is applied along the $x$ and $y$-axis, while force control governs the $z$-axis only. Hence, the rotational degrees of freedom are not implemented in this study. The parameters $K_{p}^{x}$,$ K_{d}^{x}$,$K_{p}^{f} $, and $K_{i}^{f}$ reflect the gain for PD and PI for motion and force respectively. 
Note that, PID tuning is traditionally manual or model-based, requiring continuous adaptation to varying environments. RL enhances adaptability by autonomously adjusting PID gains, improving robustness, and reducing human intervention \cite{b17}.
\subsection{Reinforcement Learning}
\begin{figure}[h!]
    \centering
    \includegraphics[width=1.0\textwidth]{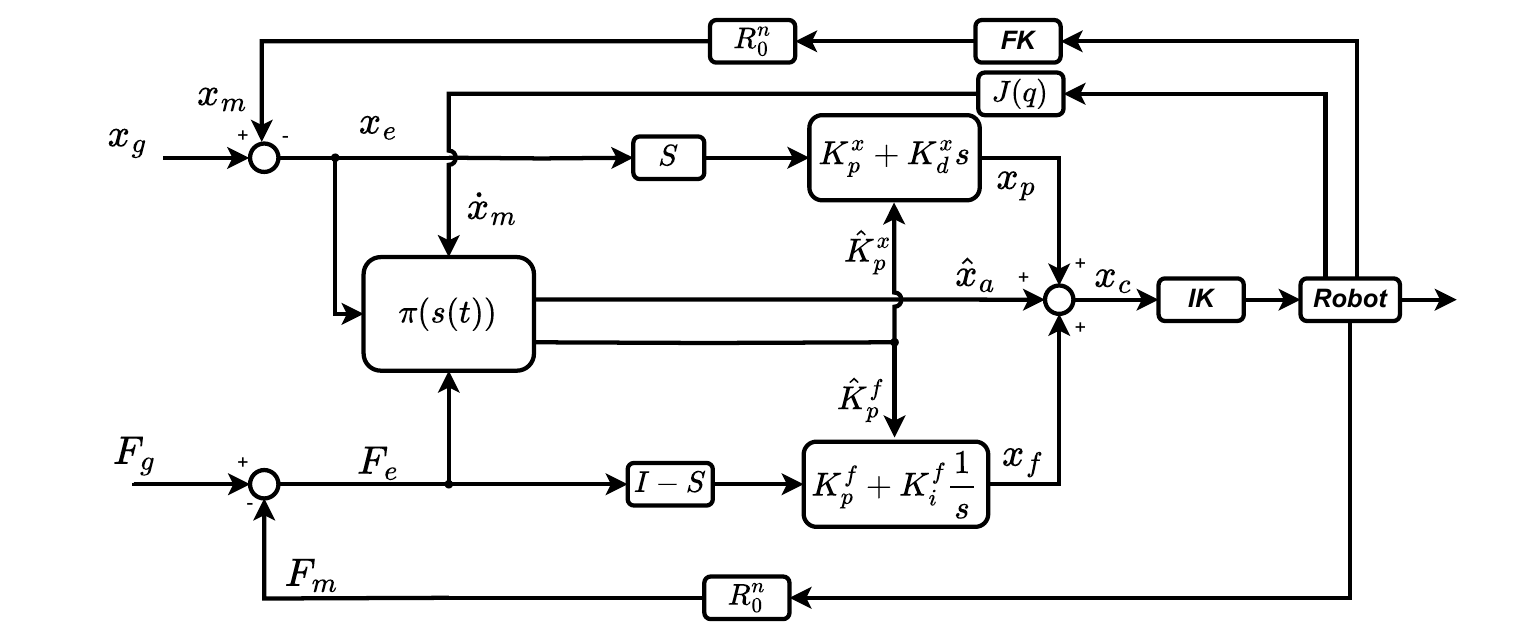}
    \caption{Diagram illustrating adaptive hybrid motion-force control. FK and IK denote forward and inverse kinematics, respectively, while $R_{0}^{n}$ defines the transformation from the respective frame into task frame.}
    \label{fig:diagramRL}
\end{figure}

RL, as a decision-making framework, enables agents to learn optimal control policy $\pi(s(t))$ through interaction with an environment. It follows trial and error principle, where an agent takes actions $a(t)$, observes states $s(t)$, rewards $r(t)$, and refines its policy to maximize long-term cumulative rewards \cite{b1}. 
\begin{figure}[hbt!]
    \centering
    \includegraphics[width=1.0\textwidth]{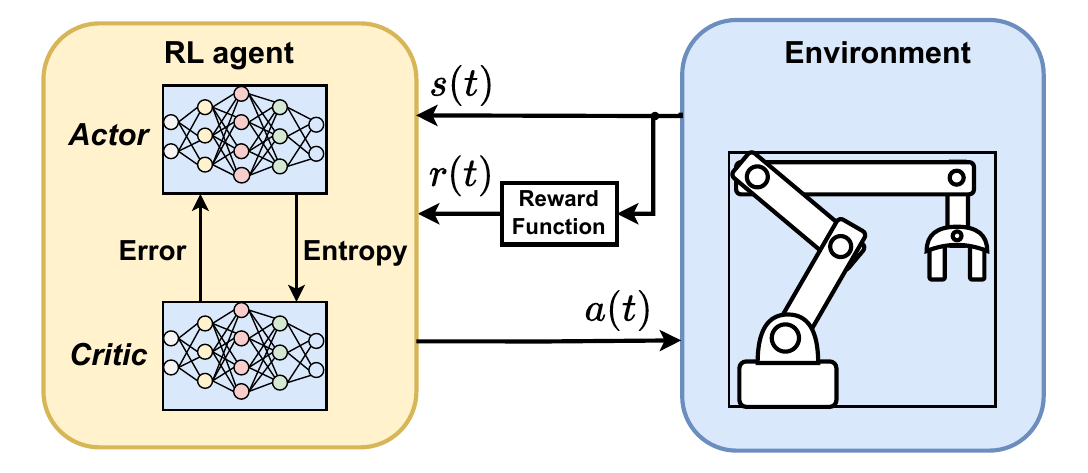}
    \caption{Diagram of the RL agent using SAC method for robot control.}
    \label{fig:diagramRL_basic}
\end{figure}  
By integrating SAC model, RL efficiently handles continuous control tasks, which is a model-free and off-policy method. It is particularly designed for continuous control tasks. The algorithm is developed upon the actor-critic framework by incorporating the maximum entropy concept, which encourages exploration and robust policy learning \cite{b19}. As illustrated in Figure \ref{fig:diagramRL}, the RL policy predicts the optimal position, $\hat{x}_a$, and the parameters of the force and motion controllers. From (\ref{eq:hybrid}), the resultant position control command, i.e. $x_c$, is defined as follows:
\begin{equation}
\begin{split}
    x_c(t) =  \hat{x}_a + u_{hybrid}(t)
    \end{split}
\end{equation}
The reward function is structured to balance long-term and short-term feedback to evaluate the behavior of the robot. The $r_{sparse}(t)$ (long-term reward) is defined as follows:
\begin{equation}
r_{sparse}(t) = \left\{ \begin{array}{rcl}
100 & \text{Task completed.} \\ 
-5  & \text{collision occurance.} \\
-5  & \text{max. time-steps occurance.}
\end{array}\right.
\end{equation}
while the $r_{dense}(t)$ (short-time reward) is computed as
\begin{equation}
    r_{dense}(t) = \alpha_1\left \| x_g - x_m \right \|_2 + \alpha_2\left \| F_g - F_{m} \right \|_2
\end{equation}
where $\left \|.\right \|_2 $ denotes to the Euclidean norm and while $r_{dense}(t)$ components are weighted via $\alpha_i$. The position and force goal, $x_g$ and $F_g$, are defined in our case-study as the hole position and $0$ force respectively.
The RL policy serves as a mapping function from the state space to the action space $\in R^{9}$, where it predicts the values of $[\hat{x}_a, K_{p}^{x}, K_{p}^{f}]$ based on the current state and reward. The remaining parameters in (\ref{eq:hybrid}) are determined as $K_{d}^{x} = 0.5 K_{p}^{x}$ and $K_{i}^{f} = 0.001 K_{p}^{f}$.
\section{Simulation Setup} \label{sec:setup}
\begin{figure}[hbt!]
\begin{subfigure}{.475\linewidth}
  \includegraphics[width=1.00\linewidth]{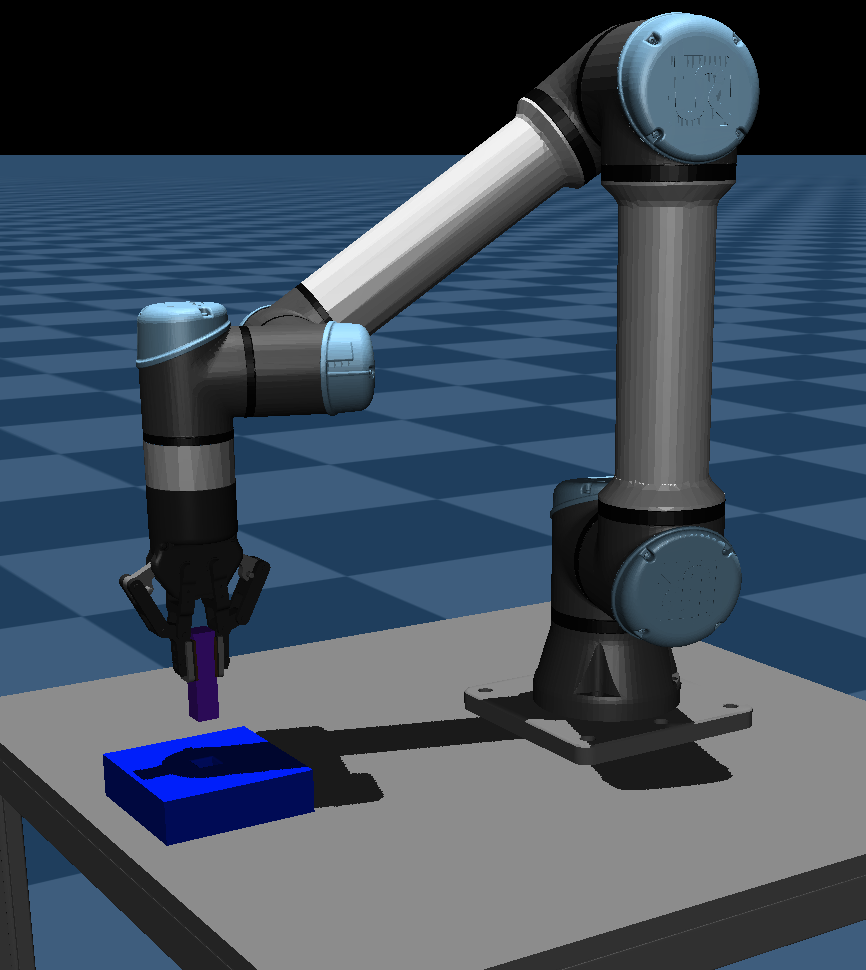}
  \caption{}
  \label{MLEDdet}
\end{subfigure}\hfill 
~ 
\begin{subfigure}{.475\linewidth}
  \includegraphics[width=0.98\linewidth]{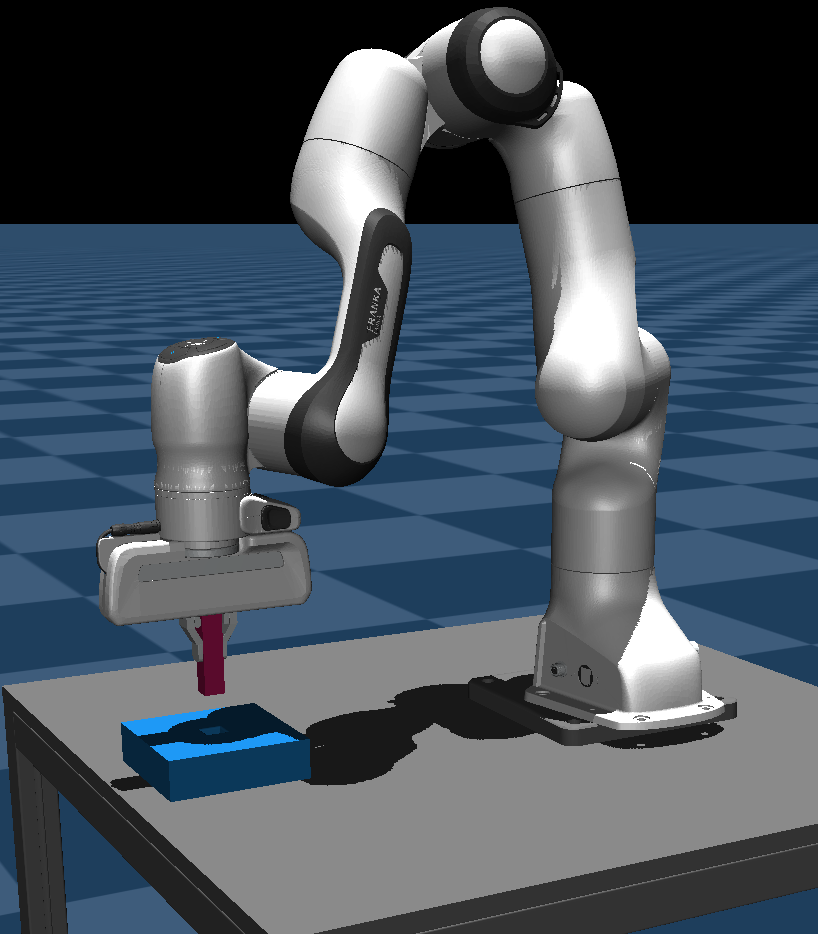}
  \caption{}
  \label{energydetPSK}
\end{subfigure}
\caption{Simulated environments of UR5e in (a) and Panda (b) models via MuJoCo.}
\label{fig:robots}
\end{figure}

In our experimental setup, the system operates on Ubuntu 20.04.6 with ROS 1 Noetic, using an AMD Ryzen 7 3700X 8 core CPU and NVIDIA GeForce RTX 3090 GPU. MuJoCo \cite{b21} is used as a simulation engine to evaluate policy transfer across robotic platforms. The UR5e and Panda robots \cite{b23}, shown in Figure \ref{fig:robots}, serve as test platforms to assess the generalizability of the learned policies. For the RL framework, we employ Stable-Baselines3 \cite{b20}. The architecture of the policy model consists of two layers of 64 neurons for the actor, and two hidden layers with 300 and 400 neurons for the critic. The control frequency of each robot is up to 60 $Hz$, while the learning frequency is 20 $Hz$. Each robot has a force/torque sensor at its wrist joint. 
This study investigates the feasibility of using UR5e as a general model for the transfer of policy between platforms. It explores how aligning the same architecture of the policy model influences domain adaptation and evaluates methods for improving skill transfer efficiency with minimal training data. 
\begin{table}[hbt!]
\centering
\caption{Proposed scenarios for evaluating policy transfer.}
\begin{tabular}{|c |c | c |} 
 \hline
 Scenario  & Use Case & Remarks \\ [0.5ex] 
 \hline
 1 & Testing UR5e model & -  \\
  \hline
 2 & Testing Panda model & -  \\
 \hline
 3 & Testing Panda using UR5e model & zero-shot   \\ 
 \hline
 4 & Testing UR5e using Panda model & zero-shot  \\ 
\hline
 5 & Testing Panda using UR5e model  & Fine-tuning\\ [1ex] 
 \hline
\end{tabular}
\label{table:scenario}
\end{table}
The agent is trained in both UR5e and Panda until the mean reward per episode stabilizes. The optimal model of the policy for both robots is saved for testing. The evaluation is conducted for each scenario listed in Table \ref{table:scenario}, where each test consists of 100 episodes with fixed and predefined starting points. This setup ensures the evaluation of the policy models, allowing for a fair comparison between different use cases.
\section{Results} \label{sec:result}
\begin{figure}[hbt!]
    \centering
    \includegraphics[width=1.0\textwidth]{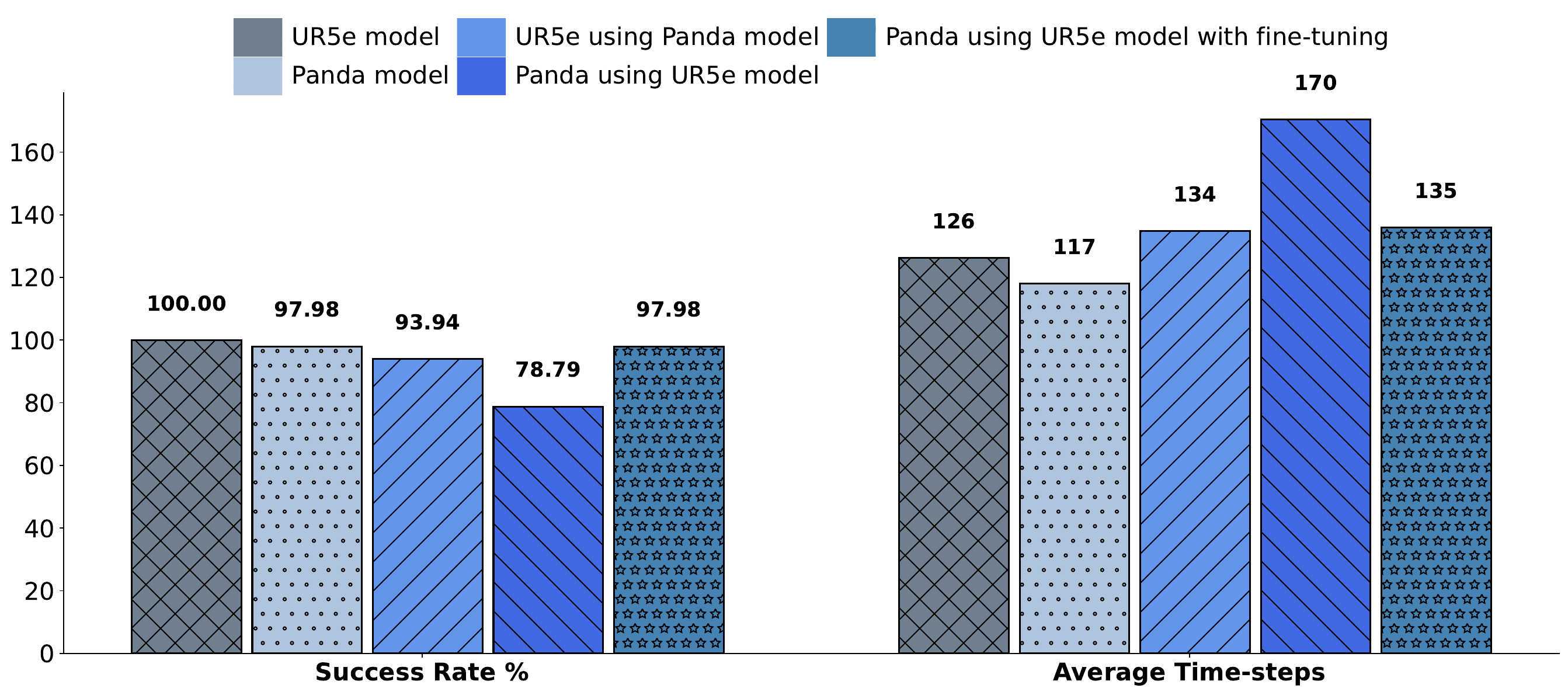}
    \caption{Results of the use-cases in Table \ref{table:scenario}.}
    \label{fig:bar_com}
\end{figure}

The section presents the results of our experiment, evaluating the performance of each robot and the effectiveness of skill transfer. The results focus on success rate and average time-steps per episode as key performance metrics, providing insights into the feasibility of using a pre-trained model from one robot to another with minimal adaptation. The success rate is computed as:
\begin{equation}
    \text{success rate \%} = \frac{\text{total num. of task completed}}{\text{total num. of episodes}} *100
\end{equation}
As illustrated in Figure \ref{fig:bar_com}, the UR5e model achieves a success rate of $100\%$, completing the task through $126$ time-steps, confirming that the policy is well-optimized when trained and tested on the same robotic platform. Similarly, the Panda model trained from scratch achieves a success rate of $97.98\%$, but with a faster average of $117$ time-steps. 
In the zero-shot transfer scenario, where the Panda model was trained using a pre-trained UR5e policy, the success rate drops to $78.79\%$, with a higher average of $170$ time-steps per episode. This suggests that while the policy can generalize to a different robotic platform, it struggles with adaptation, leading to longer execution times with less success rate. The same goes for $4^{th}$ case-scenario, UR5e performance is dropped relatively when it comes to using Panda model in a zero-shot fashion. Introducing a fine-tuning process for Panda model trained on a pre-trained UR5e policy significantly improves sample efficiency and performance. The success rate increases to $97.98\%$, with the average number of time-steps reduced to $135$, showing that adaptation techniques enhance generalization while reducing execution time in comparison with the $3^{rd}$ scenario.
\begin{figure}[hbt!]
    \centering
    \includegraphics[width=1.0\textwidth]{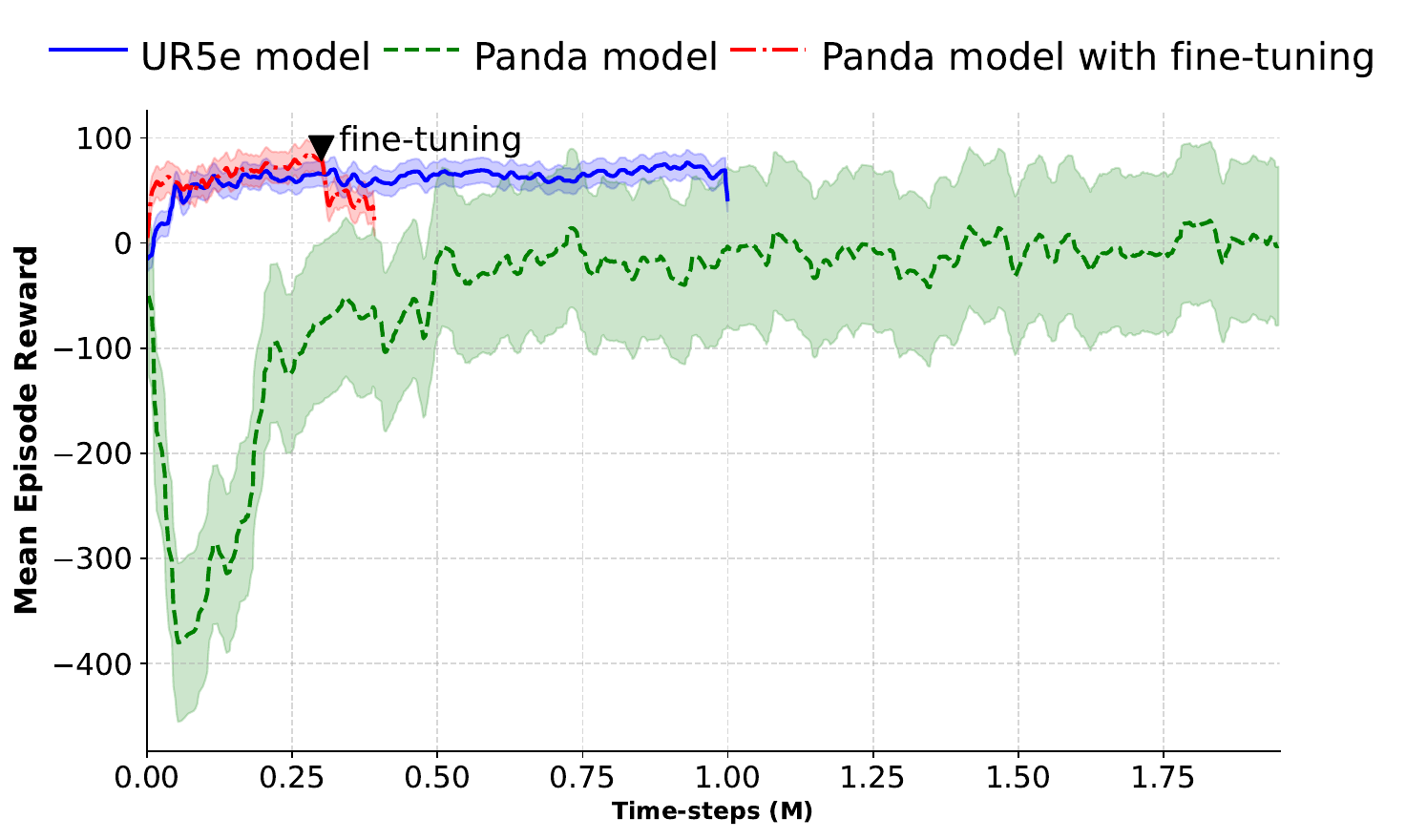}
    \caption{Comparison of the training performance for UR5e and Panda robots in MuJoCo.}
    \label{fig:training_data_com}
\end{figure}
A key observation from the training process is that UR5e converges within 200 thousand time-steps, achieving an average episode reward above $50$. In contrast, Panda model requires more than 1 million time-steps to reach a stable reward average, further confirming that training Panda model from scratch is less sample-efficient. This suggests that policy transfer from UR5e model could enhance sample efficiency, reducing the training time needed for Panda to achieve optimal performance. These results highlight that zero-shot transfer without adaptation leads to decreased performance and longer execution times, while fine-tuning significantly improves policy transfer by aligning the internal representations of different robots. The findings confirm that leveraging pre-trained models with fine-tuning is the most effective approach for improving both success rates and sample efficiency across robotic platforms.
\section{Conclusion} \label{sec:concolusion}
This study examined policy transfer in robotic manipulation, focusing on the peg-in-hole task across UR5e and Panda robots. The results demonstrate that policy transfer across robotic platforms is feasible, but adaptation mechanisms are necessary for optimal performance. While zero-shot transfer is possible, it results in lower success rates and increased execution times due to differences in configurations between platforms. In contrast, fine-tuning proves to be an effective strategy, significantly improving success rate and sample efficiency compared to training from scratch. These findings highlight the importance of adaptive policy transfer in improving sample efficiency and reducing training costs in robotic learning.
Future work will focus on transferring the UR5e policy from simulation to real-world execution with fine-tuning.
\section{Acknowledgements}
This work was funded by the Carl Zeiss Stiftung, Germany under the Sustainable Embedded AI project (P2021-02-009).

\end{document}